\documentclass[11pt]{article}

\PassOptionsToPackage{table}{xcolor}

\usepackage[final]{acl}

\usepackage{times}
\usepackage{latexsym}

\usepackage[T1]{fontenc}

\usepackage[utf8]{inputenc}

\usepackage{microtype}

\usepackage{graphicx}
\usepackage{amsmath}
\usepackage{amssymb}
\usepackage{booktabs}
\usepackage{multirow}
\usepackage{hyperref}
\usepackage{tikz}

\newcommand\citationnote{%
\footnotesize Accepted at the Main Conference of 2026 Conference on Empirical Methods in Natural Language Processing (EMNLP 2026).}
\newcommand\citationnoticebox{%
\begin{tikzpicture}[remember picture,overlay]
\node[anchor=north,yshift=-10pt] at (current page.north)
  {\fbox{\parbox{\dimexpr\textwidth-\fboxsep-\fboxrule\relax}{\citationnote}}};
\end{tikzpicture}%
}
\newcommand{\gn}[1]{{\tiny\textcolor{green!50!black}{+#1}}}
\newcommand{\rd}[1]{{\tiny\textcolor{red!60!black}{$-$#1}}}

\title{A Manifold-Aware Topic Modeling Approach via Rank-Based Prototypes}

\author{Thiago César Castilho Almeida \and Daniel Carlos Guimarães Pedronette \\
  Department of Statistics, Applied Mathematics and Computing \\
  State University of São Paulo (UNESP), Rio Claro, Brazil \\
  \texttt{\{tc.almeida, daniel.pedronette\}@unesp.br}}

\begin{document}

\maketitle
\citationnoticebox

\begin{abstract}
Recent topic models leverage pretrained embeddings, but neural architectures produce latent representations without grounding in specific texts, and clustering-based pipelines assign representative documents only post hoc, relying on absolute distances distorted by hubness and anisotropy in high-dimensional spaces. We introduce MARETopic, a training-free framework that casts topic discovery as rank-based prototype selection. After projecting embeddings onto a low-dimensional manifold, MARETopic builds ranked lists encoding ordinal neighborhood structure. A greedy algorithm selects exactly K exemplar documents, real corpus texts, whose neighborhoods cover the corpus. Two variants share this criterion. MARETopic$_\text{Corr}$ scores candidates with a query performance predictor and a rank correlation measure, leading Purity and NMI on the two benchmarks with the most categories, ahead of both neural and clustering-based topic models. MARETopic$_\text{Diff}$ scores them with a rank-based diffusion matrix, needs neither measure, and runs 1.7 to 1.9 times faster. Without a single gradient update, MARETopic leads topic coherence on two of three datasets. A novel inter-topic Maximal Marginal Relevance step raises vocabulary diversity at little cost in coherence. Our code is available at \url{https://github.com/thcastilho/maretopic}.
\end{abstract}


\section{Introduction}
\label{sec:intro}

Topic modeling aims to discover the latent thematic structure of a document collection: given $N$ documents, the goal is to identify $K$ topics---each represented by a distribution over words---and to estimate how strongly each document relates to each topic. These representations support tasks such as classification, novelty detection, corpus summarization, and similarity and relevance judgments \citep{blei2003latent}.

Two paradigms dominate the field. \emph{Generative} models represent each topic as an explicit distribution over the vocabulary, learned through probabilistic inference or optimal transport \cite{blei2003latent,dieng2020topic,wu2024fastopic}. \emph{Clustering-based} models group documents in an embedding space and extract topic words from each cluster \citep{grootendorst2022bertopic}. Despite their strengths, both paradigms leave two gaps.

The first gap is the absence of a document-level anchor for each topic. Human inspection remains central to topic model evaluation because automated coherence scores can disagree with human judgments \citep{hoyle2021automated}; yet neural models represent topics as latent vectors or distributions over the vocabulary, leaving the question \emph{``which document most represents this topic?''} without an architectural answer. Recent approaches incorporate prototypes or exemplars to improve topic learning and interpretability: ProtoXTM averages learned document representations within topic clusters as contrastive prototypes \citep{seo2025protxtm}, whereas E-LDA constructs candidate topic distributions from exemplar keywords \citep{breuer2025elda}. Neither makes an actual corpus document the semantic definition of a topic.

The second gap concerns the clustering-based paradigm. BERTopic \citep{grootendorst2022bertopic} partially addresses interpretability by exposing \emph{representative documents} per topic,\footnote{\href{https://maartengr.github.io/BERTopic/index.html\#attributes}{https://maartengr.github.io/BERTopic/index.html\#attributes}} but these are selected \emph{post-hoc} rather than being the architectural definition of the topic. Moreover, HDBSCAN \citep{campello2013hdbscan} does not take the number of topics as an input: the cluster count instead emerges from the density structure of the data, making it difficult to produce exactly $K$ topics without dataset-specific tuning or post-hoc merging.

\begin{figure}[t]
\vspace{-4mm}
    \centering
    \includegraphics[width=\columnwidth]{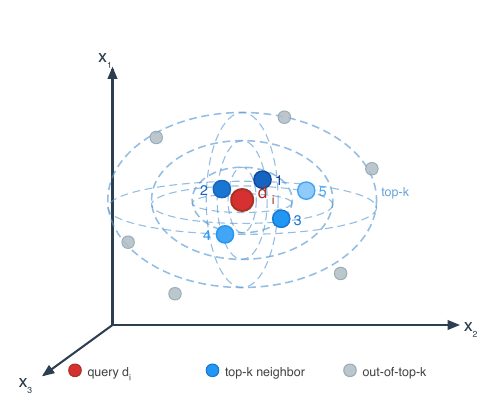}
    \vspace{-4mm}
    \caption{Spatial illustration of a ranked list for a query document $d_i$ in an embedding space. 
    Rank-based methods operate on this relative ordering rather than on absolute distances, providing robustness to the geometric distortions of high-dimensional spaces.}
    \label{fig:ranked-list-concept}
    \vspace{-3mm}
\end{figure}

Contextual analysis of neighborhoods offers a natural alternative to relying on individual pairwise scores. In high-dimensional embedding spaces, hubness causes some points to occur disproportionately often in nearest-neighbor lists \citep{radovanovic2010hubs}, while contextualized language-model representations are strongly anisotropic \citep{ethayarajh2019contextual}. Because they retain only ordinal comparisons, ranked lists are unchanged by strictly monotone transformations of the underlying distance function, a property central to ordinal embedding \citep{terada2014local}. Related rank-based similarity measures have also improved over cosine similarity in specific NLP evaluations \citep{zhelezniak2019correlations,santus2018rank}.
As illustrated in Figure \ref{fig:ranked-list-concept}, ranked lists provide a strong similarity representation structure, capable of summarizing the region surrounding a query in the embedding space. A ranked list describes an element through its context in the collection, which is what makes the representation manifold-aware: ranking is one of the families of unsupervised manifold learning \citep{pereiraferrero2024survey}, and spectral formulations make its link to the underlying manifold explicit \citep{iscen2018fastspectral}. Indeed, neighborhood information has been exploited across a range of challenging representation tasks, including the evaluation of cross-modal alignment in text–visual models \citep{Huh2024_ICML_Plato}.
In Representation Learning, rank-based methods such as RaDE \citep{fernando_rade_2022} and GRaCE \citep{almeida_grace_2025} exploit this principle to greedily select representative data points whose neighborhoods collectively cover the data manifold.

In this paper, we propose \textbf{MARETopic} (\textbf{M}anifold-\textbf{A}ware and \textbf{R}ank-based \textbf{E}xemplars Topic Modeling), a training-free topic model that operates entirely on ranked lists. Documents are encoded with Sentence-BERT \citep{reimers2019sentence}, projected onto a low-dimensional manifold via UMAP \citep{mcinnes_umap_2020}, and indexed into ranked lists. A greedy algorithm then selects $K$ exemplar documents by maximizing a scoring function while penalizing redundancy with previously selected leaders. Two complementary variants are unified under a single framework: MARETopic$_\text{Corr}$, which scores candidates via a query performance predictor (QPP) and penalizes redundancy with a rank correlation measure, and MARETopic$_\text{Diff}$, which scores via a rank-based diffusion matrix. Document--topic affinities are then converted to topic keywords. By construction, each topic is anchored on a real document in the corpus.

The main contributions of this work are:
\begin{itemize}
    \vspace{-1mm}
    \item \textit{Topic Modeling via Ranked Lists.}
    We introduce MARETopic, the first topic model built entirely on rank-based similarity information. By exploiting manifold-aware similarity over pre-trained embeddings, it discards both absolute distances and the geometric pathologies of high-dimensional spaces. Its greedy prototype selection requires no gradient updates or neural training, yielding exactly $K$ topics.
    \vspace{-1mm}
    \item \textit{Exemplar-Anchored Topic Representation.}
    MARETopic's topics are anchored in real corpus documents, directly inspectables without reference to any latent-space representation.
    \vspace{-1mm}
    \item \textit{Inter-Topic Vocabulary Diversification.}
    We introduce an inter-topic Maximal Marginal Relevance (MMR) step that penalizes vocabulary overlap among clusters, effectively increasing Topic Diversity at a minor cost in coherence.

\end{itemize}

 We evaluate MARETopic on three benchmarks (20~Newsgroups, NYT, and Web of Science) against seven state-of-the-art baselines under five metrics. The two variants rank first in seven of the fifteen dataset--metric cells, and MARETopic$_\text{Corr}$ beats every neural topic model in Purity and NMI on all three corpora without any training.

The remainder of this paper is organized as follows. Section~\ref{sec:related} surveys related work on topic modeling and rank-based methods. Section~\ref{sec:method} presents the MARETopic framework. Section~\ref{sec:experimental-evaluation} describes the experimental evaluation and presents the results, and Section \ref{sec:conclusion} concludes the work.


\section{Related Work}
\label{sec:related}

This section reviews two areas of literature foundational to our approach: advances in topic modeling architectures (Section~\ref{subsec:rw_tm}), and rank-based similarity and manifold learning (Section~\ref{subsec:rw_rank}). 

\subsection{Topic Modeling}
\label{subsec:rw_tm}

Latent Dirichlet Allocation \citep[LDA;][]{blei2003latent} and Non-negative Matrix Factorization \citep[NMF;][]{lee1999learning} laid the foundations of the field, both learning an explicit topic--word distribution---respectively by Bayesian inference and by matrix factorization of the document--term matrix.

The \textbf{generative paradigm}---models that learn an explicit topic--word distribution---has evolved through several stages of neural integration. ETM \citep{dieng2020topic} embeds words and topics in the same continuous space; CombinedTM \citep{bianchi2021pre} shows that pre-trained contextual document embeddings substantially improve topic coherence; ECRTM \citep{wu2023effective} regularizes word-topic embeddings via clustering; and FASTopic \citep{wu2024fastopic} replaces variational inference with optimal transport, achieving state-of-the-art coherence and diversity. More recently, \citet{nguyen2024contrastive} introduced setwise contrastive learning to improve latent topic representations. Across these developments, however, topics remain latent vectors or distributions over the vocabulary.

The \textbf{clustering-based paradigm}, exemplified by BERTopic \citep{grootendorst2022bertopic} and Top2Vec \citep{angelov2020top2vec}, treats topic discovery as document clustering in an embedding space. BERTopic stores \emph{representative documents} per topic, but these are selected \emph{post-hoc} by ranking a random sample of cluster members against the topic's c-TF-IDF vector; they are interpretability artifacts, not the definition of the topic itself. Moreover, HDBSCAN provides no direct control over the number of topics, and empirical studies report overproduction of topics and difficulty obtaining target counts \citep{egger2022topic}. LLM-guided clustering has been proposed as an alternative \citep{liu-etal-2025-llm-guided}, and CAST \citep{ma2025cast} improves topic word selection via corpus-aware self-similarity.

An emerging direction explores \textbf{prototype-inspired mechanisms} for topic modeling. ProtoXTM \citep{seo2025protxtm} uses document-level prototypical contrastive learning for cross-lingual alignment, while E-LDA \citep{breuer2025elda} formulates LDA topic assignment as submodular optimization over topic--document links, drawing on exemplar clustering and facility location. However, neither method uses an actual corpus document as the architectural representation of a topic.

\subsection{Ranking and Manifold Learning}
\label{subsec:rw_rank}

Standard nearest-neighbor operations rely on absolute distances in the embedding space, yet high-dimensional spaces exhibit geometric pathologies that degrade distance-based methods. \emph{Hubness}---whereby a few points appear disproportionately often in others' $k$-NN lists---is an inherent property of high-dimensional geometry \citep{radovanovic2010hubs}, confirmed in Sentence-BERT spaces \citep{nielsen2023hubness}. Concurrently, \emph{anisotropy}---the concentration of Transformer embeddings in a narrow cone---causes even unrelated representations to exhibit high cosine similarity \citep{ethayarajh2019contextual}, with subsequent work identifying isolated clusters and low-dimensional manifolds within these globally anisotropic spaces \citep{cai2021isotropy}. Ordinal embedding theory provides a principled response: \citet{terada2014local} prove that, in the large-sample limit, local ordinal comparisons suffice to recover the underlying geometry up to similarity transformations, and rank-based similarity measures have been shown to outperform cosine similarity on semantic textual similarity \citep{zhelezniak2019correlations} and outlier detection \citep{santus2018rank}. In semi-supervised NLP, mutual $k$-NN graphs---which require reciprocal neighborhood agreement---consistently outperform standard $k$-NN graphs in the evaluated document-classification task \citep{ozaki2011mutual}, supporting the robustness of reciprocal neighborhood structures.

Ranking here is more than a robustness device. \citet{pereiraferrero2024survey} surveys the field into diffusion-based, ranking-based and deep methods: a ranked list relates a query to the whole collection, so it carries neighborhood geometry that a single distance does not, and spectral formulations make the link to the underlying manifold explicit \citep{iscen2018fastspectral}. Working on ranked lists is therefore already working on the manifold. In Representation Learning, these principles have been operationalized for representative samples selection. RaDE \citep{fernando_rade_2022} constructs a log-weighted affinity matrix from ranked lists and performs diffusion to select leaders whose neighborhoods collectively cover the data manifold. GRaCE \citep{almeida_grace_2025} refines this by replacing diffusion with a query performance prediction (QPP) estimator and rank correlation measures that penalize redundant selections. Both methods produce interpretable, exemplar-anchored embeddings in which each dimension corresponds to a real data point.

Manifold learning provides a complementary perspective. Graph Laplacian regularization, incorporated into non-negative matrix factorization \citep{cai2011gnmf}, was later adopted in topic models to enforce smoothness of topic posteriors over document neighborhoods \citep{li2019lapdmm}. Graph-based models such as GTM \citep{zhou2020neural} and GNTM \citep{shen2021topic} apply neural message passing over document or word graphs.
In both cases, however, graph structure is used to inform topic distributions, not to select exemplar documents.
\citet{leticio_manifold_2024} show that neighbor-embedding projections improve image-retrieval tasks, and BERTopic \citep{grootendorst2022bertopic} uses UMAP as its default dimensionality-reduction step.

To the best of our knowledge, rank-based prototype selection has not been applied to topic modeling. MARETopic addresses this gap by adopting a greedy leader-selection strategy that operates on ranked lists, which represent the input space through ordinal neighborhood relations rather than raw Euclidean distances; projection sharpens those neighborhoods, following prior work on retrieval.


\begin{figure*}[t]
  \centering
  \vspace{-5mm}
  \includegraphics[width=0.75\linewidth]{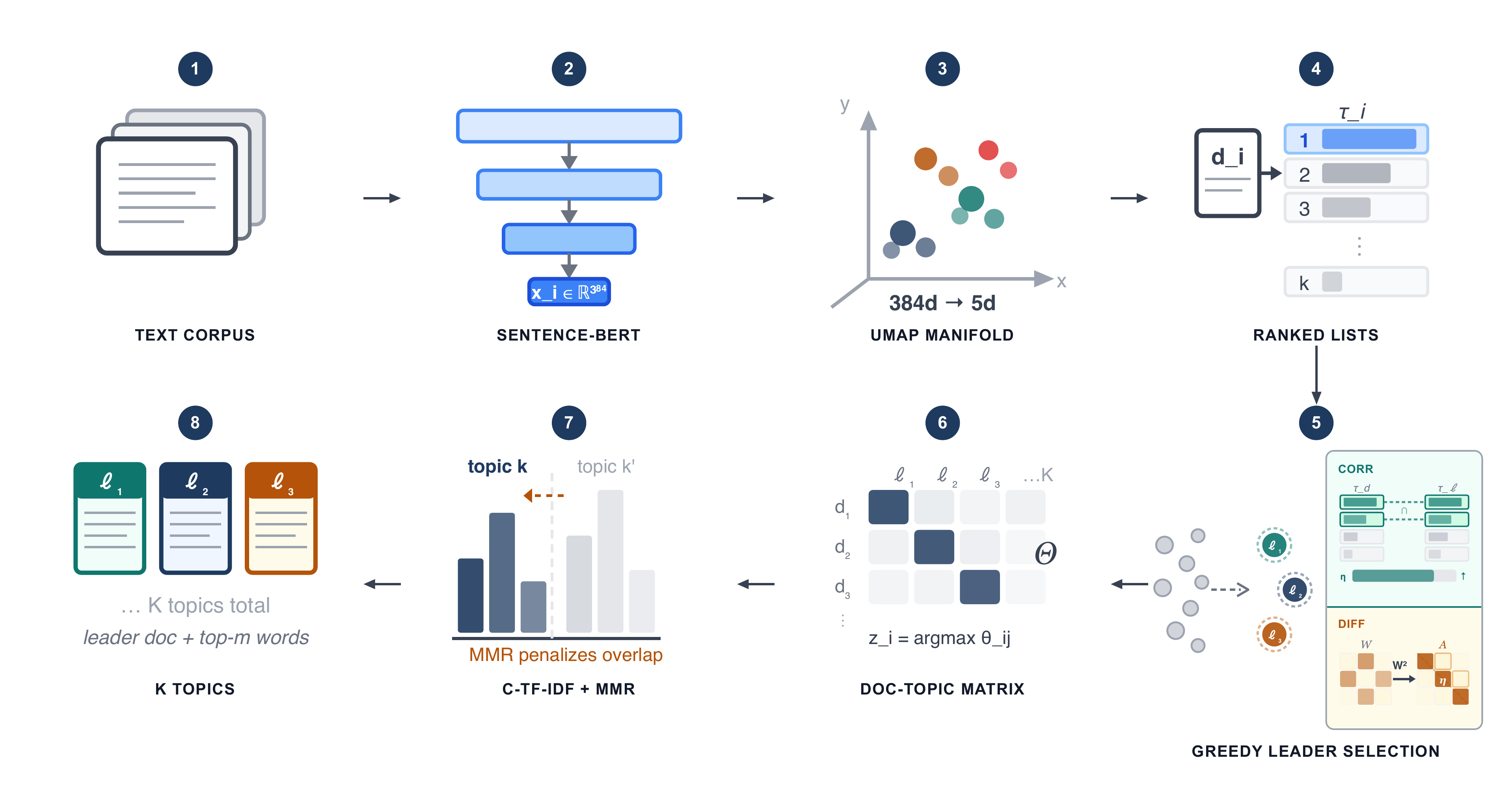}
  \vspace{-6mm}
  \caption{Overview of the MARETopic framework.}
  \label{fig:framework}
  \vspace{-2mm}
\end{figure*}

\section{The MARETopic Framework}
\label{sec:method}

Figure~\ref{fig:framework} presents an overview of the proposed MARETopic (\textbf{M}anifold-\textbf{A}ware and \textbf{R}ank-based \textbf{E}xemplars Topic Modeling) framework. 
Two variants perform greedy exemplars selection: MARETopic$_\text{Corr}$ relies on QPP and rank correlation measures, while MARETopic$_\text{Diff}$ uses a diffusion matrix. Both produce a document--topic matrix; topic words are extracted via c-TF-IDF with inter-topic MMR to yield $K$ leader documents and their representative word lists. 

\subsection{Problem Formulation}
\label{subsec:formulation}

Let $\mathcal{D} = \{d_1, d_2, \dots, d_N\}$ be a corpus of $N$ documents. Topic modeling seeks to identify $K$ latent topics, each represented by (i)~a set of characteristic words, and (ii)~a document-topic affinity matrix $\Theta \in \mathbb{R}^{N \times K}$ whose entry $\theta_{ij}$ quantifies how strongly document $d_i$ pertains to topic $j$.

MARETopic frames this as a \emph{rank-based prototype selection} problem: it identifies $K$ \emph{leader documents} $\mathcal{L} = \{\ell_1, \dots, \ell_K\} \subset \mathcal{D}$ whose neighborhoods collectively cover the corpus minimizing redundancy. Each leader anchors one topic as a real, inspectable exemplar whose neighborhood defines the topic boundary, without probabilistic modelling or neural training. The greedy selection criterion generalizes the framework of GRaCE~\cite{almeida_grace_2025} and RaDE~\cite{fernando_rade_2022} to the topic modeling setting; our contributions lie in the integration with manifold-aware text representations, the out-of-sample inference mechanism, and the inter-topic MMR diversification.

\subsection{Encoding and Indexing}
\label{subsec:encoding}

Documents in $\mathcal{D}$ are encoded with Sentence-BERT~\cite{reimers2019sentence}, producing a $d_\text{SBERT}$-dimensional feature vector $\mathbf{x}_i \in \mathbb{R}^{d_\text{SBERT}}$ per document.
Before constructing ranked lists, UMAP~\cite{mcinnes_umap_2020} projects each $\mathbf{x}_i$ from $\mathbb{R}^{d_\text{SBERT}}$ to $\mathbb{R}^{d_\text{UMAP}}$. 
The vectors are L2-normalized, and, for each document $d_i$, a traditional nearest-neighbor index~\cite{Omohundro1989balltree} is queried for the $k$ nearest neighbors (self excluded), yielding a ranked list $\tau_i$ ordering the corpus by increasing distance:
\begin{equation}
\resizebox{0.85\linewidth}{!}{$
    \displaystyle
    \tau_i(d_j) < \tau_i(d_m) \;\Longleftrightarrow\; \rho(d_i, d_j) \le \rho(d_i, d_m),
$}
\label{eq:rankedlist}
\end{equation}
where $\rho(\cdot,\cdot)$ denotes Euclidean distance and $\tau_i(d_j)$ is the rank of $d_j$ in the list of $d_i$. The top-$k$ neighborhood of $d_i$ is then:
\begin{equation}
    \mathcal{N}(d_i,k) = \{d_j \in \mathcal{D} \mid \tau_i(d_j) \le k\}.
\label{eq:neighborhood}
\end{equation}

\subsection{Rank-Based Topic Discovery}
\label{subsec:selection}

Both variants select leaders via the same greedy criterion. Let $\mathcal{L}_{t-1}$ be the leaders chosen up to step $t{-}1$. The next leader is:
\begin{equation}
    \ell_t = \arg\max_{d \,\notin\, \mathcal{L}_{t-1}}\;
    \frac{\eta(d)}{1 + \sum_{\ell \in \mathcal{L}_{t-1}} s(\tau_d, \tau_\ell)},
\label{eq:selection}
\end{equation}
where $\eta(d)$ measures the topological centrality of candidate $d$, and $s(\tau_d, \tau_\ell)$ penalizes candidates whose neighborhoods overlap with already-selected leaders to encourage topical diversity. The two variants differ only in how $\eta$ and $s$ are defined.

\subsubsection*{\textbf{MARETopic}$_\text{Corr}$}

\paragraph{Influence estimation.}
Each candidate is scored with a Query Performance Prediction (QPP) measure. In this work we use \textbf{Reciprocal Density}~\cite{pedronette_unsupervised_2015}, which measures how many neighbors of $d$ are also reciprocal neighbors of each other within $\mathcal{N}(d, k)$:
\begin{equation}
\eta(d) = \frac{1}{k^4}
\sum_{\substack{d_j \in \mathcal{N}(d,k) \\ d_i \in \mathcal{N}(d,k)}}
f_r(d_j,d_i)\;w_r(d,d_j)\;w_r(d,d_i),
\label{eq:rd}
\end{equation}
where $f_r(d_j, d_i) = |\mathcal{N}(d_j, k) \cap \{d_i\}|$ checks whether $(d_j, d_i)$ are reciprocal neighbors, and $w_r(d_j, d_i) = k + 1 - \tau_j(d_i)$ is a rank-decay weight. High values indicate documents whose neighbors agree on neighborhood membership --- a sign of locally coherent manifold structure.

\paragraph{Diversity penalization.}
Neighborhood overlap between candidate and selected leaders is measured by a rank correlation measure. Specifically, this work uses \textbf{JaccardMax}~\cite{valem_novel_2022}, which measures the overlap between two rankings by computing the maximum Jaccard coefficient across all prefix depths $l < k$:
\begin{equation}
s(\tau_d, \tau_\ell) =
    \max_{l=1}^{k-1}\;
    \frac{|\mathcal{N}(d,l) \cap \mathcal{N}(\ell,l)|}
         {|\mathcal{N}(d,l) \cup \mathcal{N}(\ell,l)|}.
\label{eq:jacmax}
\end{equation}
When JaccardMax between $d$ and a selected $\ell$ is high, their neighborhoods are largely redundant, and the denominator of Eq.~\ref{eq:selection} suppresses $d$'s score.

Any QPP estimator or rank correlation measure can replace Reciprocal Density or JaccardMax, respectively, without any changes to the other steps.

\subsubsection*{\textbf{MARETopic}$_\text{Diff}$}

\paragraph{Affinity matrix construction.}
From the ranked lists $\{\tau_i\}$, a sparse affinity matrix $\mathbf{W} \in \mathbb{R}^{N \times N}$ is built, where each entry encodes the rank position of $d_j$ in the list of $d_i$:
\vspace{-2mm}
\begin{equation}
w_{ij} =
\begin{cases}
  1 - \log_k\!\bigl(\tau_i(d_j)\bigr), & \text{if } \tau_i(d_j) \le k,\\
  0, & \text{otherwise.}
\end{cases}
\label{eq:wij}
\end{equation}

\paragraph{Diffusion scoring.}
A $p$-step diffusion $\mathbf{A} = \mathbf{W}^p$ (with $p{=}2$, following~\citet{fernando_rade_2022}) propagates local affinities across the neighborhood graph and smooths out noise in individual rank positions. The diagonal entry $a_{dd}$ captures the \emph{reciprocal affinity} of $d$ after global propagation, playing an analogous role to the QPP measure in $\text{MARETopic}_\text{Corr}$. In Eq.~\ref{eq:selection}, $\eta(d) = a_{dd}$ and $s(\tau_d, \tau_\ell) = a_{d\ell}$, penalizing candidates that share strong diffusion paths with already-selected leaders.

\subsection{Document-Topic Representation and Inference}
\label{subsec:theta}

After selecting $\mathcal{L} = \{\ell_1, \dots, \ell_K\}$, the document-topic matrix $\Theta$ is computed as:
\vspace{-2mm}
\begin{equation}
\theta_{ij} = s(\tau_i, \tau_{\ell_j}),
\qquad z_i = \arg\max_j\, \theta_{ij},
\label{eq:theta}
\end{equation}
where $s$ is the scoring function used for selection (JaccardMax for $\text{MARETopic}_\text{Corr}$; column $\ell_j$ of $\mathbf{A}$ for $\text{MARETopic}_\text{Diff}$). Hard assignment $z_i$ is used for topic-word extraction and clustering evaluation.

Out-of-sample inference reuses the fitted pipeline with no retraining: the UMAP reducer projects a new embedding into the training manifold, and the index returns its ranked list $\tau^*$ against the training corpus. For $\text{MARETopic}_\text{Corr}$, $s$ (JaccardMax) is computed directly between $\tau^*$ and each leader's list. For $\text{MARETopic}_\text{Diff}$, the diffusion matrix $\mathbf{A} = \mathbf{W}^2$ is defined only over the training graph, so a new document is instead scored by its single-step affinity $w = 1 - \log_k\!\big(\tau^*(\ell_j)\big)$ to each leader $\ell_j$ --- a one-hop approximation of the diffusion column that requires no re-diffusion.

\subsection{Topic Word Extraction}
\label{subsec:words}

Given hard assignments $\{z_i\}$, topic words are extracted in two steps. First, c-TF-IDF~\cite{grootendorst2022bertopic} scores each word $w$ for topic $c$:
\vspace{-2mm}
\begin{equation}
\resizebox{0.85\linewidth}{!}{$
    \displaystyle
    \text{c-TF-IDF}(c, w) = \text{tf}(c, w) \cdot \log\!\left(1 + \frac{\bar{n}}{\text{tf}_\mathcal{D}(w)}\right),
$}
\label{eq:ctfidf}
\end{equation}
where $\text{tf}(c, w)$ is the frequency of $w$ in the documents assigned to topic $c$, $\text{tf}_\mathcal{D}(w)$ is the total frequency of $w$ across all topics, and $\bar{n}$ is the mean number of tokens per topic.

Second, an inter-topic MMR (Maximal Marginal Relevance) step diversifies the final word lists. Topics are processed in decreasing order of cluster size. For each topic $c$, each word $w$ is re-scored as:
\begin{equation}
\resizebox{0.85\linewidth}{!}{$
    \begin{aligned}
    \text{score}(c, w) = &\;(1-\lambda)\cdot\text{c-TF-IDF}(c, w) \\
        &- \lambda\cdot \max_{c'\,\prec\, c}\;\text{c-TF-IDF}(c', w),
    \end{aligned}
$}
\label{eq:mmr}
\end{equation}
where $\lambda$ controls the diversity weight, $|c|$ denotes the cluster size, and $c' \prec c$ ranges over the topics already processed --- those with $|c'| \ge |c|$ --- so the penalty is the running maximum c-TF-IDF of $w$ over the larger topics, suppressing high-frequency words that appear in them and producing per-topic vocabularies with less lexical overlap.

Placing this step after $\Theta$ has two consequences for the evaluation. It re-ranks word lists once the assignment $z_i = \arg\max_j \theta_{ij}$ is fixed, so it cannot change that assignment: Purity and NMI do not depend on $\lambda$ at all, and only the word-level metrics do. It also never looks at the model that produced $\Theta$, so any clustering-based topic model can use it, as we verify in Appendix~\ref{app:mmrtransfer}.

\section{Experimental Evaluation}
\label{sec:experimental-evaluation}

This section evaluates MARETopic experimentally. Section~\ref{sec:setup} describes the experimental setup, including datasets, baselines, evaluation metrics, and implementation details. Section~\ref{sec:main-results} evaluates MARETopic against seven baselines across three standard benchmarks. Section~\ref{sec:cost} discuss the computational cost of the pipeline. Section~\ref{sec:ablation} investigates the sensitivity of the method to key design choices: UMAP projection, $k$ size, and MMR's $\lambda$.

\subsection{Experimental Setup}
\label{sec:setup}

\paragraph{Datasets.}
\textbf{20 Newsgroups (20NG)} \cite{Lang95twentyng} and \textbf{New York Times (NYT)}\footnote{\href{https://github.com/bobxwu/TopMost/blob/main/data/NYT.zip}{https://github.com/bobxwu/TopMost/blob/main/data/NYT.zip}} are obtained directly from the TopMost toolkit~\citep{wu2024topmost}: 20NG contains 18,846 documents across 20 topically distinct categories, and NYT comprises 9,172 articles from 12 news sections. \textbf{Web of Science (WoS)}~\citep{kowsari2017web} consists of 11,967 scientific abstracts organized into 7 disciplines,
with a stratified 80/20 train/test split.
All corpora are preprocessed by TopMost (lowercased, stopwords removed, minimum token length of three characters) and represented as bag-of-words with vocabulary sizes of 5,000 (20NG) and 10,000 (NYT and WoS), following the protocol of \citet{wu2024fastopic}.

\paragraph{Baselines.}
We compare against seven topic models spanning both paradigms: \textbf{LDA}~\citep{blei2003latent},
\textbf{NMF}~\citep{lee1999learning}, \textbf{ETM}~\citep{dieng2020topic},
\textbf{ECRTM}~\citep{wu2023effective}, and \textbf{FASTopic}~\citep{wu2024fastopic}
from the generative paradigm, and \textbf{BERTopic}~\citep{grootendorst2022bertopic}
and \textbf{Top2Vec}~\citep{angelov2020top2vec} from the clustering-based paradigm.
All methods (except Top2Vec) are evaluated through the TopMost framework~\citep{wu2024topmost} default hyperparameters (200 training epochs for neural models).

\paragraph{Evaluation Metrics.}
Five standard metrics are reported. Topic coherence is measured under the two formulations. \textbf{C$_v$} combines normalized pointwise mutual information and cosine similarity over word co-occurrence windows~\citep{roder2015exploring}; its agreement with human judgment has been questioned~\citep{hoyle2021automated}. \textbf{NPMI} normalizes pointwise mutual information to $[-1,1]$~\citep{bouma2009normalized}, with a value of zero corresponding to statistical independence for a word pair. Both are computed with \texttt{gensim}\footnote{\href{https://pypi.org/project/gensim/}{https://pypi.org/project/gensim/}} using the training corpus as the reference collection.
\textbf{TD} (Topic Diversity) measures the fraction of unique words across all topic top-word lists~\citep{dieng2020topic}. \textbf{Purity} and \textbf{NMI} assess clustering quality by comparing the hard assignment $z_i$ against ground-truth document labels.

\begin{table*}[t]
\centering
\small
\caption{Topic modeling evaluation across three datasets (mean $\pm$ std of 3 runs, $K = 50$).
Columns are grouped by paradigm. Best value per metric per dataset in \textbf{bold};
runner-up \underline{underlined}. Time is end-to-end wall clock in seconds on an Apple M4 with
16~GB of memory, averaged over three runs, and lower is better.}
\vspace{-2mm}
\label{tab:main}
\resizebox{\textwidth}{!}{%
\setlength{\tabcolsep}{3pt}
\begin{tabular}{lccccccccc}
\toprule
& \multicolumn{5}{c}{\textit{Generative}} & \multicolumn{4}{c}{\textit{Clustering-based}} \\
\cmidrule(lr){2-6}\cmidrule(lr){7-10}
& & & & & & & & \multicolumn{2}{c}{\textit{Ours}} \\
\cmidrule(l){9-10}
\textbf{Metric}
  & \textbf{LDA} & \textbf{NMF} & \textbf{ETM} & \textbf{ECRTM}
  & \textbf{FASTopic} & \textbf{BERTopic} & \textbf{Top2Vec}
  & \textbf{MARETopic} & \textbf{MARETopic} \\
& {\scriptsize\citep{blei2003latent}} & {\scriptsize\citep{lee1999learning}}
  & {\scriptsize\citep{dieng2020topic}} & {\scriptsize\citep{wu2023effective}}
  & {\scriptsize\citep{wu2024fastopic}} & {\scriptsize\citep{grootendorst2022bertopic}}
  & {\scriptsize\citep{angelov2020top2vec}}
  & {\scriptsize Diffusion} & {\scriptsize Correlation} \\
\midrule
\multicolumn{10}{l}{\textbf{20 Newsgroups}} \\[2pt]
C$_v$
  & 0.5352{\tiny$\pm$.007} & 0.5573{\tiny$\pm$.016} & 0.4522{\tiny$\pm$.004}
  & 0.6116{\tiny$\pm$.013} & 0.5993{\tiny$\pm$.019} & 0.5942{\tiny$\pm$.007}
  & 0.6649{\tiny$\pm$.011}
  & \underline{0.6708}{\tiny$\pm$.015} & \textbf{0.6818}{\tiny$\pm$.003} \\
NPMI
  & 0.0472{\tiny$\pm$.003} & 0.0613{\tiny$\pm$.007} & -0.0171{\tiny$\pm$.003}
  & -0.0372{\tiny$\pm$.021} & -0.0689{\tiny$\pm$.002} & 0.0842{\tiny$\pm$.003}
  & \textbf{0.1032}{\tiny$\pm$.003}
  & 0.0958{\tiny$\pm$.005} & \underline{0.1027}{\tiny$\pm$.002} \\
TD
  & 0.5196{\tiny$\pm$.008} & 0.5093{\tiny$\pm$.006} & 0.8471{\tiny$\pm$.005}
  & 0.8689{\tiny$\pm$.017} & \textbf{0.9849}{\tiny$\pm$.008} & 0.7733{\tiny$\pm$.002}
  & 0.8662{\tiny$\pm$.003}
  & 0.8613{\tiny$\pm$.013} & \underline{0.8787}{\tiny$\pm$.001} \\
Purity
  & 0.4204{\tiny$\pm$.022} & 0.2153{\tiny$\pm$.014} & 0.3390{\tiny$\pm$.019}
  & \underline{0.6114}{\tiny$\pm$.003} & 0.5738{\tiny$\pm$.004} & 0.4493{\tiny$\pm$.009}
  & 0.5812{\tiny$\pm$.003}
  & 0.4144{\tiny$\pm$.022} & \textbf{0.6227}{\tiny$\pm$.005} \\
NMI
  & 0.3763{\tiny$\pm$.009} & 0.1800{\tiny$\pm$.015} & 0.3032{\tiny$\pm$.018}
  & \underline{0.5348}{\tiny$\pm$.002} & 0.5202{\tiny$\pm$.005} & 0.3797{\tiny$\pm$.005}
  & 0.5265{\tiny$\pm$.004}
  & 0.4011{\tiny$\pm$.016} & \textbf{0.5419}{\tiny$\pm$.003} \\
\addlinespace
\textit{Time} (s)
  & \underline{8.6} & \textbf{5.5} & 114.3
  & 184.5 & 128.7 & 37.1
  & 52.6
  & 49.6 & 92.6 \\
\midrule
\multicolumn{10}{l}{\textbf{New York Times}} \\[2pt]
C$_v$
  & 0.4534{\tiny$\pm$.011} & 0.4919{\tiny$\pm$.010} & 0.4696{\tiny$\pm$.003}
  & 0.4457{\tiny$\pm$.023} & 0.5864{\tiny$\pm$.017} & 0.6189{\tiny$\pm$.009}
  & \textbf{0.6944}{\tiny$\pm$.013}
  & \underline{0.6570}{\tiny$\pm$.012} & 0.6496{\tiny$\pm$.018} \\
NPMI
  & 0.0002{\tiny$\pm$.005} & 0.0468{\tiny$\pm$.005} & 0.0109{\tiny$\pm$.001}
  & -0.0997{\tiny$\pm$.015} & -0.0587{\tiny$\pm$.005} & \underline{0.1166}{\tiny$\pm$.002}
  & \textbf{0.1295}{\tiny$\pm$.006}
  & 0.1079{\tiny$\pm$.006} & 0.1060{\tiny$\pm$.004} \\
TD
  & 0.5769{\tiny$\pm$.022} & 0.4560{\tiny$\pm$.011} & 0.8427{\tiny$\pm$.009}
  & \underline{0.9595}{\tiny$\pm$.008} & \textbf{0.9996}{\tiny$\pm$.001} & 0.7951{\tiny$\pm$.004}
  & 0.9071{\tiny$\pm$.008}
  & 0.8831{\tiny$\pm$.006} & 0.8702{\tiny$\pm$.002} \\
Purity
  & 0.5454{\tiny$\pm$.024} & 0.4612{\tiny$\pm$.003} & 0.5556{\tiny$\pm$.011}
  & 0.6529{\tiny$\pm$.002} & 0.6583{\tiny$\pm$.018} & 0.6184{\tiny$\pm$.006}
  & \underline{0.6874}{\tiny$\pm$.006}
  & 0.6046{\tiny$\pm$.016} & \textbf{0.7084}{\tiny$\pm$.003} \\
NMI
  & 0.2661{\tiny$\pm$.020} & 0.1952{\tiny$\pm$.007} & 0.2815{\tiny$\pm$.005}
  & 0.3549{\tiny$\pm$.004} & 0.3670{\tiny$\pm$.012} & 0.3221{\tiny$\pm$.006}
  & \underline{0.3890}{\tiny$\pm$.002}
  & 0.3305{\tiny$\pm$.011} & \textbf{0.3975}{\tiny$\pm$.002} \\
\addlinespace
\textit{Time} (s)
  & \underline{7.2} & \textbf{4.5} & 148.2
  & 223.7 & 123.9 & 36.9
  & 44.2
  & 31.6 & 53.6 \\
\midrule
\multicolumn{10}{l}{\textbf{Web of Science}} \\[2pt]
C$_v$
  & 0.4797{\tiny$\pm$.006} & 0.4869{\tiny$\pm$.014} & 0.4198{\tiny$\pm$.006}
  & 0.4655{\tiny$\pm$.004} & 0.6000{\tiny$\pm$.005} & 0.6920{\tiny$\pm$.005}
  & 0.7018{\tiny$\pm$.011}
  & \textbf{0.7198}{\tiny$\pm$.004} & \underline{0.7150}{\tiny$\pm$.005} \\
NPMI
  & -0.0287{\tiny$\pm$.011} & 0.0396{\tiny$\pm$.006} & -0.0045{\tiny$\pm$.005}
  & -0.1076{\tiny$\pm$.001} & -0.0308{\tiny$\pm$.009} & 0.1244{\tiny$\pm$.007}
  & 0.1344{\tiny$\pm$.003}
  & \underline{0.1359}{\tiny$\pm$.003} & \textbf{0.1447}{\tiny$\pm$.002} \\
TD
  & 0.6636{\tiny$\pm$.017} & 0.4876{\tiny$\pm$.007} & 0.8938{\tiny$\pm$.010}
  & \underline{0.9996}{\tiny$\pm$.001} & \textbf{1.0000}{\tiny$\pm$.000} & 0.7782{\tiny$\pm$.010}
  & 0.8796{\tiny$\pm$.012}
  & 0.8569{\tiny$\pm$.013} & 0.8547{\tiny$\pm$.001} \\
Purity
  & 0.6529{\tiny$\pm$.009} & 0.6086{\tiny$\pm$.012} & 0.6456{\tiny$\pm$.008}
  & 0.7733{\tiny$\pm$.010} & 0.7765{\tiny$\pm$.012} & 0.7934{\tiny$\pm$.004}
  & \textbf{0.8293}{\tiny$\pm$.009}
  & 0.6082{\tiny$\pm$.021} & \underline{0.8106}{\tiny$\pm$.003} \\
NMI
  & 0.3875{\tiny$\pm$.005} & 0.2962{\tiny$\pm$.010} & 0.3853{\tiny$\pm$.011}
  & 0.4518{\tiny$\pm$.008} & 0.4565{\tiny$\pm$.005} & 0.4473{\tiny$\pm$.003}
  & \textbf{0.4901}{\tiny$\pm$.003}
  & 0.3660{\tiny$\pm$.008} & \underline{0.4664}{\tiny$\pm$.002} \\
\addlinespace
\textit{Time} (s)
  & \underline{6.5} & \textbf{5.2} & 170.7
  & 253.7 & 137.2 & 35.0
  & 49.6
  & 37.7 & 65.2 \\
\bottomrule
\end{tabular}%
}
\end{table*}

\paragraph{Implementation Details.}
All experiments use $K = 50$ topics and three independent runs, following \citep{grootendorst2022bertopic}, reporting means and additionally standard deviations ($\pm$). All methods are evaluated under the same number of topics to ensure a controlled comparison; Appendix~\ref{app:k100} repeats the whole evaluation at $K = 100$. The default configuration for both variants sets the neighborhood size to $k = 100$, the UMAP target dimensionality to $d_\text{UMAP} = 5$ (with $n_\text{neighbors} = 15$, $\text{min\_dist} = 0.0$, cosine metric, mirroring BERTopic's defaults), and the MMR diversity parameter to $\lambda = 0.3$. Each topic is represented by $m = 15$ top words extracted via c-TF-IDF followed by inter-topic MMR diversification. All methods that operate on document embeddings --- FASTopic, BERTopic, MARETopic$_\text{Corr}$, and MARETopic$_\text{Diff}$ --- use the \texttt{all-MiniLM-L6-v2} SBERT encoder, ensuring that differences in topic quality reflect the modeling mechanism rather than the encoder. Robustness to this choice is verified in Appendix~\ref{app:encoder}, where three SBERT backbones yield consistent results.

\paragraph{Code Availability.}
The implementation of MARETopic with full integration with the TopMost framework~\citep{wu2024topmost}, is available at \href{https://github.com/thcastilho/maretopic}{https://github.com/thcastilho/maretopic}.


\subsection{Main Results}
\label{sec:main-results}

Table~\ref{tab:main} presents the results across all three datasets and five metrics. Qualitative results are provided in Appendices~\ref{app:topics} and \ref{app:heatmap}.

\paragraph{Clustering quality.}
MARETopic$_\text{Corr}$, which requires no gradient updates, achieves the highest Purity and NMI on 20NG and NYT --- the two corpora with the most categories --- surpassing neural models trained for 200 epochs. The largest margin is on NYT, where it gains $+5.0$~pp in Purity over FASTopic and $+2.1$~pp over Top2Vec, the strongest clustering-based baseline. On WoS the ordering reverses and Top2Vec leads both metrics, a boundary we return to in Section~\ref{sec:conclusion}. MARETopic$_\text{Diff}$ trails the correlation variant in clustering throughout; on 20NG its Purity falls below several baselines, reflecting that the diffusion scoring separates fine-grained categories less sharply. Its strengths lie instead in coherence.

\paragraph{Topic coherence.}
Under C$_v$ the two variants are most complementary: MARETopic$_\text{Corr}$ achieves the highest value on 20NG ($0.6818$, ahead of the nearest baseline Top2Vec by $1.7$~pp) and MARETopic$_\text{Diff}$ the highest on WoS ($0.7198$, ahead of Top2Vec by $1.8$~pp), with the other variant immediately behind in both cases; on NYT, Top2Vec leads at $0.6944$, $3.7$~pp above MARETopic$_\text{Diff}$.

NPMI orders the methods differently, dividing them by paradigm. Every generative neural model scores non-positive NPMI on at least two corpora, FASTopic and ECRTM on all three: the words they group into a topic co-occur no more often than chance would predict, which is exactly what a globally fitted distribution risks. Every clustering-based method stays well above zero. Inside that group the margins are small: MARETopic$_\text{Corr}$ leads on WoS ($0.1447$), sits within $0.0005$ of Top2Vec on 20NG, inside the run-to-run deviation, and falls behind clustering-based baselines on NYT.

\paragraph{Topic diversity.}
Diversity follows paradigm lines: FASTopic's Sinkhorn regularization enforces near-perfect TD by construction, a structural advantage that no clustering-based method currently matches. Within the clustering-based paradigm, however, the proposed inter-topic MMR step effectively mitigates the vocabulary overlap inherent to these approaches: with the default penalty, both variants cleanly exceed BERTopic in TD ($0.85$--$0.88$ vs.\ $0.77$--$0.80$). Top2Vec occupies the same band and edges the two variants on NYT and WoS ($0.9071$ and $0.8796$).

\subsection{Computational Cost}
\label{sec:cost}

MARETopic trains nothing, so it pays no epoch cost, and the Time row of Table~\ref{tab:main} reflects that. The figures cover encoding, projection, indexing, topic discovery, out-of-sample inference and word extraction. MARETopic$_\text{Diff}$ sits in the same band as the other embedding-based methods, while MARETopic$_\text{Corr}$ costs $1.7$--$1.9\times$ the diffusion variant, since accumulating a JaccardMax affinity over the neighborhood is more expensive than reading a diffusion diagonal. Both variants undercut every neural topic model on every corpus, where the cost goes to 200 training epochs.

Indexing, the step a rank-based method adds to the pipeline, is not where the cost lies. Building the exact nearest-neighbor index takes $0.18$--$0.41$~s across the three corpora, under $1\%$ of end-to-end runtime in every configuration; the time is dominated by document encoding and by leader selection instead. Selection uses only rank positions, never the distances behind them, so the exact index can be swapped for an approximate one at sublinear query cost without changing the method.

\subsection{Ablation Studies}
\label{sec:ablation}

Sections~\ref{sec:ablation-umap}--\ref{sec:ablation-mmr} analyze sensitivity to three internal design choices of MARETopic: the UMAP pre-processing step, the neighborhood size $k$, and the MMR diversity parameter $\lambda$. All ablations follow the same protocol from Section~\ref{sec:setup}.\footnote{Ablation experiments are conducted as independent evaluation runs from the main comparison in Table~\ref{tab:main}; minor numerical deviations are within the observed standard deviation and reflect UMAP stochasticity.} Each study reports C$_v$, TD, Purity and NMI, except where a design choice cannot affect one of them.

\subsubsection{Effect of UMAP Pre-processing}
\label{sec:ablation-umap}

\begin{table}[t]
\centering
\small
\caption{Effect of UMAP pre-processing.
Each cell shows the UMAP-enabled value with the delta from raw SBERT
({\color{green!50!black}green}: improvement; {\color{red!60!black}red}: degradation).}
\label{tab:umap}
\setlength{\tabcolsep}{3pt}
\begin{tabular}{lccc}
\toprule
& \textbf{20NG} & \textbf{NYT} & \textbf{WoS} \\
\midrule
\multicolumn{4}{l}{\textbf{MARETopic$_\text{Corr}$}} \\[2pt]
C$_v$  & 0.6817\,\rd{0.026} & 0.6449\,\rd{0.053} & 0.7189\,\rd{0.028} \\
TD     & 0.8854\,\gn{0.025} & 0.8764\,\gn{0.016} & 0.8542\,\rd{0.014} \\
Purity & 0.6161\,\gn{0.017} & 0.7102\,\gn{0.061} & 0.8094\,\rd{0.002} \\
NMI    & 0.5398\,\gn{0.031} & 0.3982\,\gn{0.041} & 0.4697\,\gn{0.002} \\
\midrule
\multicolumn{4}{l}{\textbf{MARETopic$_\text{Diff}$}} \\[2pt]
C$_v$  & 0.6750\,\gn{0.039} & 0.6458\,\rd{0.044} & 0.7177\,\rd{0.009} \\
TD     & 0.8787\,\rd{0.019} & 0.8831\,\gn{0.011} & 0.8756\,\rd{0.002} \\
Purity & 0.4103\,\rd{0.004} & 0.5908\,\rd{0.027} & 0.6089\,\rd{0.097} \\
NMI    & 0.3984\,\gn{0.020} & 0.3241\,\rd{0.002} & 0.3623\,\rd{0.044} \\
\bottomrule
\end{tabular}
\end{table}

\begin{figure*}[t]
\vspace{-4mm}
\centering
\includegraphics[width=.7\textwidth]{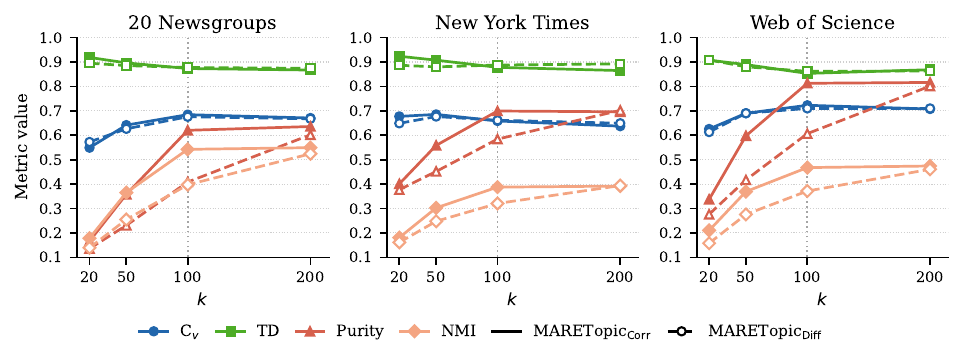}
\vspace{-3mm}
\caption{Sensitivity to neighborhood size $k$ across all four metrics.
Solid lines: MARETopic$_\text{Corr}$; dashed lines: MARETopic$_\text{Diff}$.
Vertical dotted line marks the recommended $k=100$.}
\label{fig:topk}
\end{figure*}

\begin{table*}[t]
\centering
\caption{Sensitivity to the MMR diversity parameter $\lambda$.
Only C$_v$ and TD are reported, as MMR acts on word extraction and does not affect cluster assignments.
Colored deltas show change relative to the previous $\lambda$
({\color{green!50!black}green}: improvement; {\color{red!60!black}red}: degradation).
Recommended $\lambda = 0.3$ highlighted (gray).}
\label{tab:mmr}
\resizebox{0.8\textwidth}{!}{%
\setlength{\tabcolsep}{4pt}
\begin{tabular}{llcccccc}
\toprule
& & \multicolumn{2}{c}{\textbf{20 Newsgroups}} & \multicolumn{2}{c}{\textbf{New York Times}} & \multicolumn{2}{c}{\textbf{Web of Science}} \\
\cmidrule(lr){3-4}\cmidrule(lr){5-6}\cmidrule(lr){7-8}
\textbf{Method} & $\lambda$ & C$_v$ & TD & C$_v$ & TD & C$_v$ & TD \\
\midrule
  & 0.0
    & \textbf{0.6777} & 0.6627
    & \textbf{0.6398} & 0.6845
    & \textbf{0.7299} & 0.6938 \\
\rowcolor{gray!15}
  & 0.3
    & 0.6674\,\rd{0.010} & 0.8813\,\gn{0.219}
    & 0.6384\,\rd{0.001} & 0.8729\,\gn{0.188}
    & 0.7206\,\rd{0.009} & 0.8600\,\gn{0.166} \\
  & 0.5
    & 0.6539\,\rd{0.014} & 0.9538\,\gn{0.073}
    & 0.6349\,\rd{0.004} & 0.9582\,\gn{0.085}
    & 0.6850\,\rd{0.036} & 0.9533\,\gn{0.093} \\
\multirow{-4}{*}{\shortstack{\textbf{MARETopic}\\{\scriptsize Correlation}}}
  & 0.7
    & 0.6257\,\rd{0.028} & \textbf{0.9925}\,\gn{0.039}
    & 0.6043\,\rd{0.031} & \textbf{0.9893}\,\gn{0.031}
    & 0.6689\,\rd{0.016} & \textbf{0.9902}\,\gn{0.037} \\
\midrule
  & 0.0
    & \textbf{0.6854} & 0.6502
    & 0.6516 & 0.7080
    & \textbf{0.7287} & 0.6906 \\
\rowcolor{gray!15}
  & 0.3
    & 0.6676\,\rd{0.018} & 0.8800\,\gn{0.230}
    & \textbf{0.6677}\,\gn{0.016} & 0.8747\,\gn{0.167}
    & 0.7080\,\rd{0.021} & 0.8569\,\gn{0.166} \\
  & 0.5
    & 0.6487\,\rd{0.019} & 0.9564\,\gn{0.076}
    & 0.6506\,\rd{0.017} & 0.9569\,\gn{0.082}
    & 0.6816\,\rd{0.026} & 0.9582\,\gn{0.101} \\
\multirow{-4}{*}{\shortstack{\textbf{MARETopic}\\{\scriptsize Diffusion}}}
  & 0.7
    & 0.6322\,\rd{0.017} & \textbf{0.9920}\,\gn{0.036}
    & 0.6322\,\rd{0.018} & \textbf{0.9893}\,\gn{0.032}
    & 0.6577\,\rd{0.024} & \textbf{0.9938}\,\gn{0.036} \\
\bottomrule
\end{tabular}%
}
\end{table*}

Table~\ref{tab:umap} isolates the contribution of UMAP by reporting each metric alongside its delta from raw SBERT embeddings, and the two variants respond differently. For MARETopic$_\text{Corr}$, the projection consistently improves clustering on 20NG and NYT at the cost of lower C$_v$, because the compression partly disrupts vocabulary co-occurrence patterns. For MARETopic$_\text{Diff}$ it improves C$_v$ and NMI on 20NG but degrades Purity on NYT and substantially on WoS, where the seven broad categories are already well-separated in raw space, and the projection introduces unnecessary distortion.

This asymmetry reflects a fundamental difference between the two scoring mechanisms. The QPP and rank correlation measures in MARETopic$_\text{Corr}$ rely on fine-grained neighborhood structure and benefit from the tighter local neighborhoods that UMAP provides. The diffusion operator in MARETopic$_\text{Diff}$, by contrast, already smooths local noise through matrix multiplication and gains less from the projection. We retain $d_\text{UMAP}=5$ as the default because it yields consistent clustering gains for MARETopic$_\text{Corr}$ on 20NG and NYT; for MARETopic$_\text{Diff}$ or coarser-grained corpora, disabling UMAP is a reasonable choice.

\subsubsection{Neighborhood Size ($k$)}
\label{sec:ablation-topk}

Figure~\ref{fig:topk} shows the effect of $k$ on all four metrics. Purity and NMI rise steeply from $k=20$ to $k=100$ for both variants. MARETopic$_\text{Corr}$ levels off near $k=100$, while MARETopic$_\text{Diff}$ keeps gaining beyond it, because the diffusion operator benefits from denser neighborhoods --- on 20NG and WoS Purity its $100 \rightarrow 200$ gain still exceeds the previous step. The word-level metrics are far less sensitive to $k$: for $k \ge 50$, both C$_v$ (peaking around $k=50$--$100$) and TD (declining gently) vary by at most $0.06$, the only pronounced movement being the rise in C$_v$ from $k=20$ to $k=50$. We set $k=100$ as the default: it is the elbow of the clustering curves for MARETopic$_\text{Corr}$, and the cost of that variant grows with $k$, since JaccardMax scans every prefix $l < k$.

\subsubsection{MMR Diversity Parameter ($\lambda$)}
\label{sec:ablation-mmr}

MMR $\lambda$ is varied to map the coherence--diversity trade-off. Table~\ref{tab:mmr} reports C$_v$ and TD for $\lambda \in \{0.0, 0.3, 0.5, 0.7\}$. TD is sensitive to $\lambda$: without MMR, TD falls below BERTopic because exemplar-based clusters share much of their vocabulary. At $\lambda = 0.3$, TD rises by roughly $+0.2$ on every dataset for both variants, while C$_v$ drops by at most $0.021$; we adopt this as the default. At $\lambda = 0.7$, TD nearly matches FASTopic ($\approx 0.99$) but C$_v$ decreases more noticeably. The step acts on word lists after $\Theta$ is fixed, so Purity and NMI do not move with $\lambda$ and are omitted from the table. Nor is the step specific to MARETopic: Appendix~\ref{app:mmrtransfer} attaches it to BERTopic's own c-TF-IDF, where it raises Topic Diversity by a comparable margin.

\section{Conclusion}
\label{sec:conclusion}

This work introduced MARETopic, a training-free framework that reframes topic modeling as rank-based prototype selection. By operating on ranked-list neighborhoods over a low-dimensional manifold, MARETopic sidesteps the hubness and anisotropy pathologies that distort absolute distances in high-dimensional embedding spaces, and inherently anchors each topic to a real, inspectable corpus document. Coupled with a novel inter-topic Maximal Marginal Relevance (MMR) extraction step, the framework ranks first in seven of the fifteen dataset--metric cells, and MARETopic$_\text{Corr}$ beats every neural topic model in Purity and NMI on all three corpora without any training. The diversity gap to optimal-transport models such as FASTopic belongs to the clustering-based paradigm rather than to this method, and the MMR step that narrows it works on any competitor. On Web of Science, the most separable of the three corpora, Top2Vec reaches higher Purity and NMI; where the manifold is already simple, a centroid and an exemplar may describe the same region, which puts the advantage of rank-based selection on more irregular collections. Future work will extend this rank-based formulation to hierarchical topic structures and dynamic topic modeling over time, complement the evaluation with downstream and intrusion-based protocols, and assess scalability on larger and multilingual corpora.

\section*{Limitations}

Our evaluation covers three English-language corpora of moderate size (9K--19K documents) with 7--20 categories. Although these are standard benchmarks in topic modeling~\citep{wu2024fastopic,wu2024topmost}, the effectiveness on multilingual or domain-specific corpora (e.g., legal, biomedical) remains to be verified. Section~\ref{sec:cost} measures the nearest-neighbor index that MARETopic adds and finds its cost negligible, but the exact index still scales with corpus size, and very large collections (100K+ documents) would need an approximate one. We have not evaluated that regime. The runtime comparison is also limited by our hardware: the neural baselines have no GPU path here and run CPU-bound. What we claim is architectural, that MARETopic needs neither training nor an accelerator. We do not claim to be faster in wall-clock terms.
Topic Diversity remains below generative models such as FASTopic, a gap inherent to the clustering-based paradigm rather than specific to our method; the inter-topic MMR step that narrows it is method-agnostic (Appendix~\ref{app:mmrtransfer}). Finally, a downstream evaluation would strengthen the practical impact claim, measuring the impact of the proposal in tasks such as document classification and word/topic intrusion.

\section*{Acknowledgments}

The authors are grateful to the National Council for Scientific and Technological Development—CNPq (grant \#313193/2023-1), the São Paulo Research Foundation—FAPESP (grant \#2024/04890-5), and Petrobras (grant \#2023/00095-3) for their financial support.

\bibliography{references}

\appendix

\section{Transferability of the Inter-Topic MMR}
\label{app:mmrtransfer}

\begin{table*}[t]
\centering
\small
\caption{The proposed diversification applied to BERTopic's own c-TF-IDF (mean of 3 runs, $K = 50$). Deltas are against
BERTopic's native word lists ({\color{green!50!black}green}: increase;
{\color{red!60!black}red}: decrease). At $\lambda = 0$ the step reduces to selecting the
highest-scoring words, which is what BERTopic does natively, and reproduces its word lists
exactly on every run. Purity and NMI are unchanged by
construction, since the diversification acts on word lists after $\Theta$ is fixed. This is an independent run, so BERTopic's native
values differ slightly from those in Table~\ref{tab:main}.}
\label{tab:mmrtransfer}
\resizebox{\textwidth}{!}{%
\setlength{\tabcolsep}{4pt}
\begin{tabular}{lcccccccccccc}
\toprule
& \multicolumn{4}{c}{\textbf{20 Newsgroups}}
& \multicolumn{4}{c}{\textbf{New York Times}}
& \multicolumn{4}{c}{\textbf{Web of Science}} \\
\cmidrule(lr){2-5}\cmidrule(lr){6-9}\cmidrule(lr){10-13}
\textbf{Configuration}
  & C$_v$ & TD & Purity & NMI
  & C$_v$ & TD & Purity & NMI
  & C$_v$ & TD & Purity & NMI \\
\midrule
BERTopic, native words
  & 0.6120 & 0.7666 & 0.4261 & 0.3770
  & 0.6341 & 0.8067 & 0.6187 & 0.3155
  & 0.6903 & 0.7707 & 0.7909 & 0.4488 \\
\quad + inter-topic MMR, $\lambda = 0.3$
  & 0.5991\,\rd{0.013} & 0.9382\,\gn{0.172} & 0.4261 & 0.3770
  & 0.6445\,\gn{0.010} & 0.9382\,\gn{0.132} & 0.6187 & 0.3155
  & 0.6531\,\rd{0.037} & 0.9195\,\gn{0.149} & 0.7909 & 0.4488 \\
\bottomrule
\end{tabular}%
}
\end{table*}

This section evaluates whether the inter-topic MMR step is specific to MARETopic, by attaching it to a competing model. The step re-ranks word lists without looking at the model that produced them, so any clustering-based topic model can use it. Table~\ref{tab:mmrtransfer} applies it to the c-TF-IDF scores of BERTopic, leaving that method's clustering, vocabulary and scoring formula untouched, so word selection is the only thing that changes. Topic Diversity rises by $0.132$--$0.172$, carrying BERTopic from the $0.77$--$0.81$ of its native word lists to $0.92$--$0.94$, at a cost of at most $0.037$ in C$_v$. On NYT, C$_v$ improves by $0.010$ instead. That is above the $0.85$--$0.88$ MARETopic itself reaches on the same corpora. 
Diversification is therefore a swappable component rather than an advantage of MARETopic.

The same experiment answers a second question: how much of the reported advantage comes from the rank-based selection of leaders, and how much from the machinery MARETopic and BERTopic share, since both extract topic words with c-TF-IDF. Giving BERTopic the same word pipeline MARETopic uses, c-TF-IDF followed by inter-topic MMR at $\lambda = 0.3$, MARETopic$_\text{Corr}$ still reaches a higher C$_v$ on all three corpora: $0.6818$ against BERTopic's $0.5991$ on 20NG, and $0.7150$ against $0.6531$ on WoS. On NYT the margin is narrow, $0.6496$ against $0.6445$, well inside the standard deviation of either measurement, so the two are better read as tied. 

\section{Number of Topics}
\label{app:k100}

\begin{table*}[t]
\centering
\small
\caption{The evaluation repeated at $K = 100$ topics (mean of 3 runs). Each cell carries the value at $K = 100$ and, in small type, its change from $K = 50$ ({\color{green!50!black}green}: increase; {\color{red!60!black}red}: decrease). Best value per metric per dataset in \textbf{bold}; runner-up \underline{underlined}.}
\label{tab:k100}
\resizebox{\textwidth}{!}{%
\setlength{\tabcolsep}{2pt}
\begin{tabular}{lccccccccc}
\toprule
& \multicolumn{5}{c}{\textit{Generative}} & \multicolumn{4}{c}{\textit{Clustering-based}} \\
\cmidrule(lr){2-6}\cmidrule(lr){7-10}
& & & & & & & & \multicolumn{2}{c}{\textit{Ours}} \\
\cmidrule(l){9-10}
\textbf{Metric}
  & \textbf{LDA} & \textbf{NMF} & \textbf{ETM} & \textbf{ECRTM}
  & \textbf{FASTopic} & \textbf{BERTopic} & \textbf{Top2Vec}
  & \textbf{MARETopic} & \textbf{MARETopic} \\
& & & & & & &
  & {\scriptsize Diffusion} & {\scriptsize Correlation} \\
\midrule
\multicolumn{10}{l}{\textbf{20 Newsgroups}} \\[2pt]
C$_v$
  & 0.4817\,\rd{0.053} & 0.5444\,\rd{0.013} & 0.4508\,\rd{0.001} & 0.5751\,\rd{0.037} & 0.5584\,\rd{0.041} & 0.5874\,\rd{0.007} & 0.5995\,\rd{0.065} & \underline{0.6020}\,\rd{0.069} & \textbf{0.6123}\,\rd{0.070} \\
TD
  & 0.5678\,\gn{0.048} & 0.4129\,\rd{0.096} & 0.8111\,\rd{0.036} & 0.8098\,\rd{0.059} & \textbf{0.9160}\,\rd{0.069} & 0.7218\,\rd{0.051} & \underline{0.8548}\,\rd{0.011} & 0.8413\,\rd{0.020} & 0.8138\,\rd{0.065} \\
Purity
  & 0.4220\,\gn{0.002} & 0.2222\,\gn{0.007} & 0.3724\,\gn{0.033} & 0.6111\,{\tiny 0.000} & \underline{0.6202}\,\gn{0.046} & 0.4916\,\gn{0.042} & 0.6123\,\gn{0.031} & 0.5875\,\gn{0.173} & \textbf{0.6537}\,\gn{0.031} \\
NMI
  & 0.3527\,\rd{0.024} & 0.1785\,\rd{0.002} & 0.3133\,\gn{0.010} & 0.5149\,\rd{0.020} & 0.5178\,\rd{0.002} & 0.4042\,\gn{0.025} & \textbf{0.5362}\,\gn{0.010} & 0.4878\,\gn{0.087} & \underline{0.5269}\,\rd{0.015} \\
\midrule
\multicolumn{10}{l}{\textbf{New York Times}} \\[2pt]
C$_v$
  & 0.4327\,\rd{0.021} & 0.4874\,\rd{0.005} & 0.4418\,\rd{0.028} & 0.4853\,\gn{0.040} & 0.5492\,\rd{0.037} & \underline{0.6651}\,\gn{0.046} & \textbf{0.7001}\,\gn{0.006} & 0.6432\,\rd{0.014} & 0.6244\,\rd{0.025} \\
TD
  & 0.6320\,\gn{0.055} & 0.3920\,\rd{0.064} & 0.8025\,\rd{0.040} & \underline{0.9087}\,\rd{0.051} & \textbf{0.9911}\,\rd{0.009} & 0.7389\,\rd{0.056} & 0.8822\,\rd{0.025} & 0.8422\,\rd{0.041} & 0.8293\,\rd{0.041} \\
Purity
  & 0.5744\,\gn{0.029} & 0.5254\,\gn{0.064} & 0.6086\,\gn{0.053} & 0.6924\,\gn{0.039} & 0.6950\,\gn{0.037} & 0.6594\,\gn{0.041} & 0.6964\,\gn{0.009} & \underline{0.7273}\,\gn{0.123} & \textbf{0.7280}\,\gn{0.020} \\
NMI
  & 0.2897\,\gn{0.024} & 0.2409\,\gn{0.046} & 0.3223\,\gn{0.041} & 0.3728\,\gn{0.018} & 0.3796\,\gn{0.013} & 0.3432\,\gn{0.021} & 0.3858\,\rd{0.003} & \underline{0.3926}\,\gn{0.062} & \textbf{0.3929}\,\rd{0.005} \\
\midrule
\multicolumn{10}{l}{\textbf{Web of Science}} \\[2pt]
C$_v$
  & 0.4572\,\rd{0.023} & 0.4436\,\rd{0.043} & 0.3756\,\rd{0.044} & 0.4992\,\gn{0.034} & 0.5412\,\rd{0.059} & \underline{0.6447}\,\rd{0.047} & \textbf{0.6568}\,\rd{0.045} & 0.6218\,\rd{0.098} & 0.6359\,\rd{0.079} \\
TD
  & 0.7020\,\gn{0.038} & 0.4142\,\rd{0.073} & 0.8902\,\rd{0.004} & \underline{0.9829}\,\rd{0.017} & \textbf{0.9940}\,\rd{0.006} & 0.7044\,\rd{0.074} & 0.8703\,\rd{0.009} & 0.8340\,\rd{0.023} & 0.8156\,\rd{0.039} \\
Purity
  & 0.6929\,\gn{0.040} & 0.6041\,\rd{0.005} & 0.6043\,\rd{0.041} & 0.7831\,\gn{0.010} & 0.7971\,\gn{0.021} & 0.7927\,\rd{0.001} & \textbf{0.8352}\,\gn{0.006} & 0.8140\,\gn{0.206} & \underline{0.8336}\,\gn{0.023} \\
NMI
  & 0.3772\,\rd{0.010} & 0.2854\,\rd{0.011} & 0.3036\,\rd{0.082} & 0.4293\,\rd{0.022} & 0.4336\,\rd{0.023} & 0.4176\,\rd{0.030} & \textbf{0.4837}\,\rd{0.006} & 0.4364\,\gn{0.070} & \underline{0.4474}\,\rd{0.019} \\
\bottomrule
\end{tabular}%
}
\end{table*}

This section repeats the evaluation at a larger topic budget to assess whether the quality of greedy selection degrades as $K$ grows. All results in Section~\ref{sec:main-results} fix $K = 50$, the standard setting under which the baselines are published. Because MARETopic selects its topics greedily, raising $K$ extends a sequence it has already begun rather than refitting from scratch: later leaders are chosen from whatever earlier ones left uncovered, and could in principle be weaker. Table~\ref{tab:k100} reports the outcome at $K = 100$. Among the methods that reach that budget, MARETopic$_\text{Corr}$ still leads Purity and NMI on all three corpora, with MARETopic$_\text{Diff}$ as runner-up on NYT and WoS. Doubling the budget, therefore, does not erode the advantage.

Top2Vec is the one method that cannot be held to the budget: its topic count follows from the density structure of the embedding space, and its reduction step merges topics but never splits them, so a request for $100$ returns $84$, $80$ and $71$ topics. Its numbers are therefore not measured at the same budget as the others. On NYT the two MARETopic variants take first and second place in both Purity and NMI, ahead of Top2Vec on each.

The deltas show what raising $K$ changes, and little of it is specific to MARETopic. C$_v$ falls for 23 of the 27 method--corpus pairs and TD for 24, as expected when a fixed vocabulary is divided among twice as many topics, while Purity rises for all but four. The three largest Purity gains belong to MARETopic$_\text{Diff}$ ($+0.21$ on WoS, $+0.17$ on 20NG, $+0.12$ on NYT), which closes most of its gap to the correlation variant: its weakness at $K = 50$ is a matter of resolution rather than of the diffusion scoring itself.

\section{Encoder Robustness}
\label{app:encoder}

\begin{table*}[t]
\centering
\small
\caption{Robustness to SBERT backbone (all-MiniLM-L6-v2, all-mpnet-base-v2, all-distilroberta-v1),
with all other parameters fixed at default ($k=100$, $d_\text{UMAP}=5$, $\lambda=0.3$, $K=50$,
mean of 3 runs). Best value per method per dataset per metric in \textbf{bold}.}
\label{tab:encoder}
\resizebox{\textwidth}{!}{%
\setlength{\tabcolsep}{4pt}
\begin{tabular}{llcccccccccccc}
\toprule
& & \multicolumn{4}{c}{\textbf{20 Newsgroups}}
& \multicolumn{4}{c}{\textbf{New York Times}}
& \multicolumn{4}{c}{\textbf{Web of Science}} \\
\cmidrule(lr){3-6}\cmidrule(lr){7-10}\cmidrule(lr){11-14}
\textbf{Method} & \textbf{Encoder}
  & C$_v$ & TD & Purity & NMI
  & C$_v$ & TD & Purity & NMI
  & C$_v$ & TD & Purity & NMI \\
\midrule
\multirow{3}{*}{\shortstack{\textbf{MARETopic}\\{\scriptsize Correlation}}}
  & MiniLM        & 0.6818 & \textbf{0.8787} & \textbf{0.6227} & \textbf{0.5419}
                  & 0.6496 & 0.8702 & \textbf{0.7084} & \textbf{0.3975}
                  & \textbf{0.7150} & 0.8547 & 0.8106 & 0.4664 \\
  & MPNet         & \textbf{0.6875} & 0.8653 & 0.6141 & 0.5405
                  & \textbf{0.6633} & \textbf{0.8840} & 0.7062 & 0.3950
                  & 0.7033 & \textbf{0.8715} & 0.8019 & 0.4647 \\
  & DistilRoBERTa & 0.6686 & 0.8609 & 0.6170 & 0.5326
                  & 0.6310 & 0.8658 & 0.6997 & 0.3874
                  & 0.7097 & 0.8591 & \textbf{0.8158} & \textbf{0.4753} \\
\midrule
\multirow{3}{*}{\shortstack{\textbf{MARETopic}\\{\scriptsize Diffusion}}}
  & MiniLM        & 0.6708 & 0.8613 & \textbf{0.4144} & \textbf{0.4011}
                  & 0.6570 & \textbf{0.8831} & 0.6046 & 0.3305
                  & \textbf{0.7198} & 0.8569 & 0.6082 & 0.3660 \\
  & MPNet         & \textbf{0.6754} & \textbf{0.8707} & 0.4040 & 0.3932
                  & 0.6684 & 0.8649 & \textbf{0.6078} & \textbf{0.3341}
                  & 0.6995 & \textbf{0.8800} & 0.5894 & 0.3687 \\
  & DistilRoBERTa & 0.6567 & 0.8582 & 0.3837 & 0.3752
                  & \textbf{0.6752} & 0.8711 & 0.5930 & 0.3266
                  & 0.7127 & 0.8605 & \textbf{0.6328} & \textbf{0.3838} \\
\bottomrule
\end{tabular}%
}
\end{table*}

\begin{figure*}[t]
  \centering
  \includegraphics[width=\linewidth]{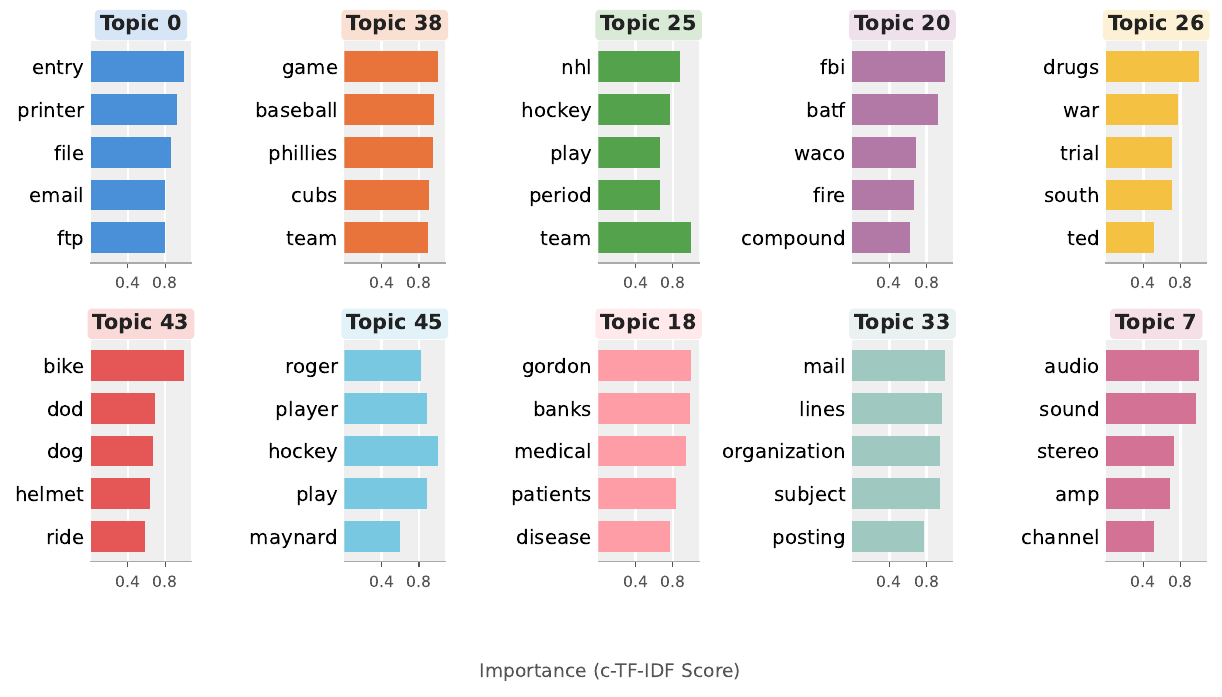}
  \caption{Ten representative topics from MARETopic$_\text{Corr}$ on 20NG. Each panel is a leader document; bars show its top-5 c-TF-IDF terms. Topic indices follow the ordering from the greedy selection.}
  \label{fig:topics}
\end{figure*}

This section evaluates the sensitivity of both MARETopic variants to the choice of SBERT backbone. Table~\ref{tab:encoder} reports all four metrics across three encoders with all other parameters fixed at default ($k=100$, $d_\text{UMAP}=5$, $\lambda=0.3$, $K=50$, mean of 3 runs). Both methods are robust to encoder choice: for MARETopic$_\text{Corr}$, Purity varies by at most 1.4~pp and NMI by at most 1.1~pp across encoders; MARETopic$_\text{Diff}$ is also stable, though its spreads are wider: up to 4.3~pp in Purity and 2.6~pp in NMI. Topic diversity is slightly more sensitive (up to 2.3~pp), which is expected given c-TF-IDF's dependence on the vocabulary structure of each embedding space; still, all encoders keep TD in the 0.85--0.88 range. This stability is a consequence of operating on relative neighborhood order rather than absolute distances: as long as the embedding spaces are of comparable quality, the ranked lists --- and therefore the selected leaders --- remain largely the same. No single encoder dominates across all settings; MiniLM is retained as default for consistency with the BERTopic baseline.

\section{Qualitative Topic Examples}
\label{app:topics}

This section presents qualitative examples of the topics discovered by MARETopic$_\text{Corr}$. Figure~\ref{fig:topics} shows ten representative topics on 20NG. The topics cover diverse domains such as computing, sports, law enforcement, vehicles, and medicine, with highly concise keywords and little vocabulary overlap, demonstrating the quality of the greedy leader-selection.

\section{Document--Topic Matrix Visualization}
\label{app:heatmap}

\begin{figure*}[t]
  \centering
  \includegraphics[width=\linewidth]{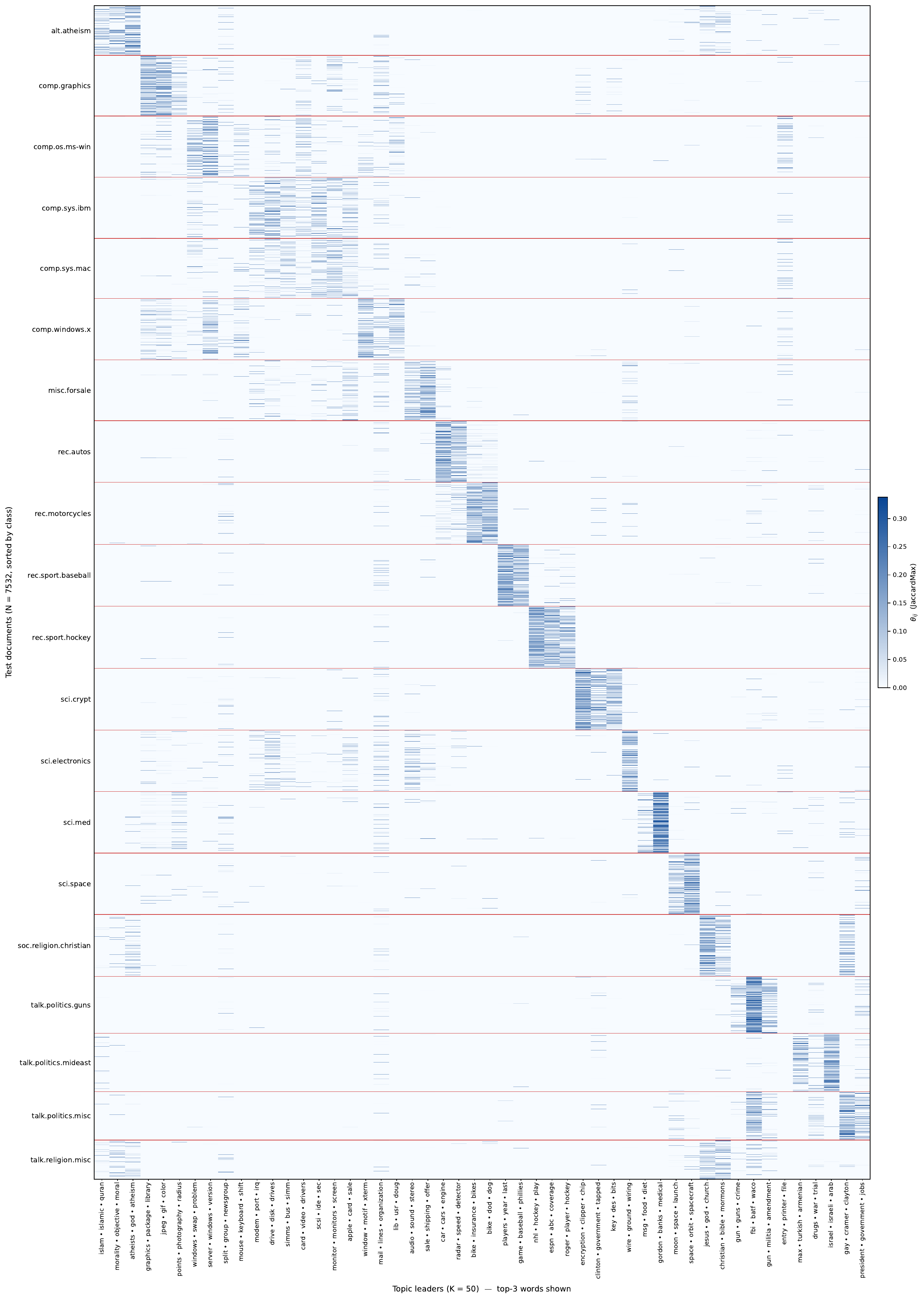}
  \caption{Document--topic affinity matrix $\Theta$ for MARETopic$_\text{Corr}$
    on the 20NG test set ($N{=}7{,}532$, $K{=}50$).
    Rows: topic leaders sorted by dominant class, labelled by top-3 words.
    Columns: test documents sorted by ground-truth class; red lines mark
    class boundaries. Color: $\theta_{ij}$ (JaccardMax affinity).}
  \label{fig:heatmap}
\end{figure*}

This section visualizes the document--topic affinity matrix to assess whether the rank-based assignment generalizes to unseen data. Figure~\ref{fig:heatmap} shows $\Theta$ for all 7,532 20NG test documents, computed via a single \texttt{transform()} call with no retraining. The block-diagonal structure emerging on held-out data confirms that the rank-based affinity generalises beyond the training corpus: each leader attracts documents from its corresponding class while remaining near-zero elsewhere.

\end{document}